\documentclass{article}
\usepackage{ICASSP2022,amsmath,amssymb,graphicx}
\usepackage[table]{xcolor}
\usepackage{hyperref}
\hypersetup{hidelinks}
\usepackage[ruled]{algorithm2e}
\usepackage{booktabs,cite}
\usepackage[super]{nth}
\usepackage{booktabs}
\usepackage{multirow}
\usepackage{url}
\usepackage{bbding}       

\definecolor{Gray}{gray}{0.9}

\title{Effectiveness of Text, Acoustic, and Lattice-based representations in Spoken Language Understanding tasks}
\name{
\shortstack{
Esaú Villatoro-Tello$^{\star, 1}$, 
Srikanth Madikeri$^{\star,1}$, 
Juan Zuluaga-Gomez$^{1,2}$,
Bidisha Sharma$^{5}$, \\
\textit{Seyyed Saeed Sarfjoo}$^{1}$,
\textit{Iuliia Nigmatulina}$^{1,3}$,
\textit{Petr Motlicek}$^{1,4}$,
\textit{Alexei V. Ivanov}$^{5}$,
\textit{Aravind Ganapathiraju}$^{5}$
\thanks{$^{\star}$Corresponding authors: \{\textit{esau.villatoro, srikanth.madikeri}\}@idiap.ch}
\thanks{This work was supported by the Idiap \& Uniphore collaboration project.}
}
}

\address{
 $^{1}$ Idiap Research Institute, Martigny, Switzerland.\\
 $^{2}$ Ecole Polytechnique Federale de Lausanne (EPFL), Switzerland.\\
 $^{3}$ Institute of Computational Linguistics, University of Zürich, Switzerland.\\
 $^{4}$Brno University of Technology, Brno, Czech Republic\\
 $^{5}$ Uniphore Software Systems Inc., Palo Alto, CA, USA. 
}

\begin{document}
\ninept

\maketitle

\begin{abstract}
In this paper, we perform an exhaustive evaluation of different representations to address the intent classification problem in a Spoken Language Understanding (SLU) setup. We benchmark three types of systems to perform the SLU intent detection task: 1) text-based, 2) lattice-based, and a novel 3) multimodal approach. Our work provides a comprehensive analysis of what could be the achievable performance of different state-of-the-art SLU systems under different circumstances, e.g., automatically- \textit{vs.} manually-generated transcripts. We evaluate the systems on the publicly available SLURP spoken language resource corpus. 
Our results indicate that using richer forms of Automatic Speech Recognition (ASR) outputs, namely word-consensus-networks, allows the SLU system to improve in comparison to the 1-best setup (5.5\% relative improvement). However, crossmodal approaches, i.e., learning from acoustic and text embeddings, obtains performance similar to the oracle setup, a relative improvement of 17.8\% over the 1-best configuration, being a recommended alternative to overcome the limitations of working with automatically generated transcripts. 
\end{abstract}



\section{Introduction}
\label{sec-Intro}
Spoken Language Understanding (SLU) is the underlying key component of interactive smart devices such as voice assistants, social bots, and intelligent home devices. Effectively interpreting human interactions through classification of intent and slot filling plays a crucial role in SLU. Therefore, it is not surprising that the SLU problem has received substantial attention in industry and academia.

Typically, SLU aims at parsing spoken utterances into corresponding structured semantic concepts through a pipeline approach. First, the spoken utterances are transcribed by an automatic speech recognition (ASR) system, and the transcribed audio is subsequently processed by a natural language understanding (NLU) module to identify the intent and extract slots from the utterance.\footnote{In practice, the 1-best transcript representation is the one sent to the natural language understanding (NLU) model for intent detection.}  The main disadvantages of the pipeline approach include: (1) errors in the ASR transcripts are directly propagated to the NLU module, normally trained only on correct transcriptions; (2) prosodical and non-phonetic aspects present in the spoken utterance are not taken into account. Even though, the classical text-based approach is mostly used in industrial applications and is still an active research area~\cite{simonnet2017asr}.

More recently, end-to-end (E2E) SLU systems have gained popularity~\cite{serdyuk2018towards,ref14_bidisha,haghani2018audio}. E2E SLU acts as an individual single model, and it directly predicts the intent from speech without exploiting an intermediate text representation. In particular, it directly optimizes the performance metrics of SLU. Due to the complex structure of speech signals, a large SLU database along with high-end computational resources are required for training E2E models. In~\cite{haghani2018audio}, several E2E SLU encoder-decoder solutions are investigated. For instance, instead of directly mapping speech to SLU target~\cite{serdyuk2018towards}, pre-trained acoustic and language models can be used for downstream SLU tasks, showing to be an effective paradigm~\cite{denisov2020pretrained,Lugosch2019}. Similarly, there have been efforts to design tighter integration of ASR and NLU systems beyond 1-best ASR results, e.g., by means of encoding several ASR hypotheses through lattice-based representations. A lattice is a compact representation encoding multiple ASR hypotheses obtained at the decoding step. Its use has shown to be key in boosting the performance of IR systems \cite{VillatoroSIGIR2022}. In this direction, there are several works adopting word confusion networks (WCNs) as input to NLU systems to preserve information in possible hypotheses~\cite{tur2013semantic, Liu2020wcn, zou2021lattice-slu}. The main advantage of WCN-based approaches is that they are less sensitive to the ASR errors. Finally, recent approaches based on multi-modal information have been proposed ~\cite{ref14_bidisha}. The main motivation behind this idea is founded on how humans interpret, in the real world, the meaning of an utterance and corresponding semantics from various cues, thus, assuming that the acoustic and linguistic content of a speech signal may carry complementary information for deriving robust semantic information of an utterance.

\begin{figure*}[t]
    \centering
    \includegraphics[width=0.96\linewidth]{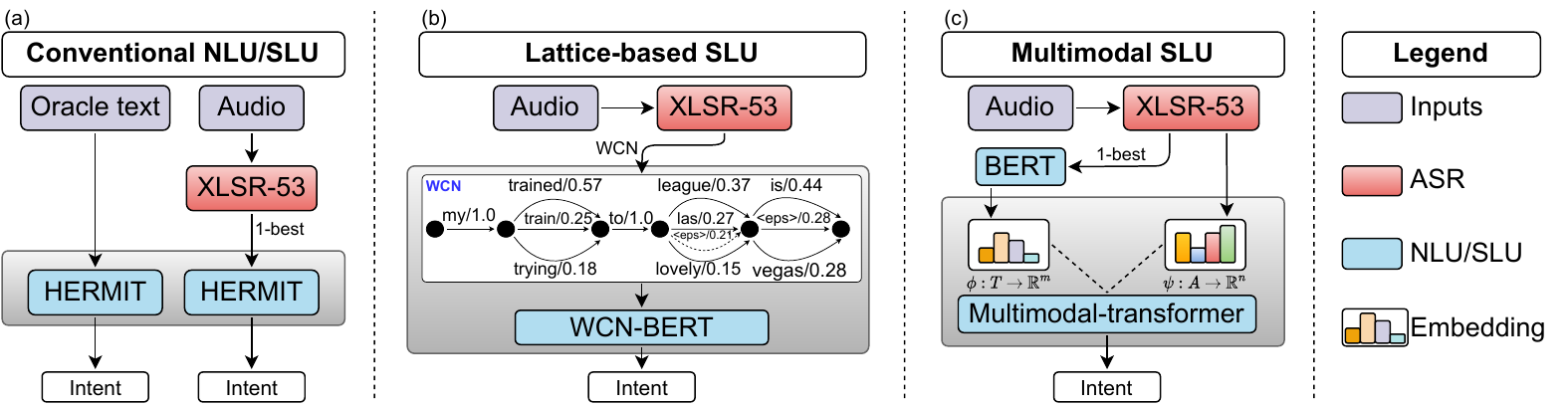}
    \caption{Overview of the considered NLU/SLU methodologies for our performed experiments.}
    \label{fig:experiments}
\end{figure*}

Overall, despite the promising results, there still exists a gap between the demonstrated capability of SLU systems and the requirements of an industrial application, e.g.,  a generalized voice assistant. For instance, E2E approaches mostly focus on databases with limited semantic complexity and structural diversity~\cite{mckenna2020semantic}. Additionally, most of the current benchmarks on SLU are widely saturated, where the obtained performances (F1-scores) are near perfect. Examples of such cases are results reported on ATIS~\cite{hemphill1990atis}, Fluent Speech Commands~\cite{Lugosch2019}, or SNIPS~\cite{coucke2018snips} datasets. Hence, in order to validate the robustness of recent SLU approaches under a more realistic scenario, it becomes necessary to focus on SLU tasks that incorporate more complex semantics and numerous intent classes and slots. To the best of our knowledge, such benchmarking, comparing the wide variety of SLU approaches, has not been performed recently.  

In this paper, we present an extensive analysis of different SLU techniques, ranging from pure text-based alternatives to methods that are able to process richer forms of ASR outputs. Overall, our work has three salient features: \textit{(1)} we evaluate and compare under the same circumstances four big families of SLU approaches, namely: text-based, lattice-based, multimodal, and end-to-end, \textit{(2)} our performed evaluation is done considering a more realistic scenario, i.e., where no access to manual transcriptions exists, but instead, ASR transcripts are given as input to the SLU systems, and \textit{(3)} we describe several inconsistencies found in the SLURP~\cite{bastianelli2020slurp} dataset, one of the most challenging test beds for SLU systems.\footnote{Our code is publicly available: \url{https://github.com/idiap/slu_representations}}

\section{Methodology}

Figure \ref{fig:experiments} depicts an overview of the considered SLU techniques in our experiments: (a) conventional or pipeline-oriented NLU/SLU approaches, (b) Lattice-based SLU architectures, and (c) multimodal (text+acoustic) architectures. Although not shown in the figure, we also report the performance of very recent E2E methods.

\subsection{Conventional NLU/SLU systems}
\label{subsec:pipeline_slu}
We selected the HERMIT architecture~\cite{vanzo-etal-2019-hierarchical} as the representative approach for this category of systems.  HERMIT, a \textbf{H}i\textbf{ER}archical \textbf{M}ult\textbf{I}-\textbf{T}ask Natural Language Understanding architecture, was designed for effective semantic parsing of domain-independent user utterances, extracting meaning representations in terms of high-level intents and frame-like semantic structures. According to the authors, HERMIT stands out for being a cross-domain, multi-task architecture, capable of recognizing multiple intents in human-machine interactions. The central motivation behind the design of the HERMIT architecture is the modeling of the dependence among three tasks, namely, dialogue acts identification, intents, and slots. For this, the authors addressed the NLU problem using a seq2seq model employing BiLSTM encoders and self-attention mechanisms, followed by
CRF tagging layers. HERMIT was validated in two large datasets with a high number of intent labels (58 to 68 classes), reporting a performance of F1=86\%. 

We re-implemented HERMIT in PyTorch~\cite{pytorch2019}, with the following changes: we exchanged the encoder layer based on ELMO embeddings with a BERT \cite{devlin2019bert} encoder, we replace the BiLTSM encoders by GRU modules and used the AdamW optimizer. We evaluate the performance of our implementation of HERMIT when either, manual transcriptions (1-best) ASR outputs (extracted from the XLS-R model, see Section \ref{subsect:xlsr}) are given as inputs (see Figure \ref{fig:experiments}.a).  

\subsection{Lattice-based SLU}
\label{subsec:WCN}

As described earlier, one main limitation of pipeline SLU systems is their sensitivity to the errors present in the ASR transcriptions. Consequently, there have been proposed several approaches for building SLU systems robust against ASR errors based on lattices and WCNs~\cite{Liu2020wcn, zou2021lattice-slu}. Although both, word lattices and WCNs contain more information than N-best lists, WCNs have been proven more efficient in terms of size and structure, thus representing a more plausible alternative when designing SLU systems that receive as input a graph-based structure. 

We re-implemented a very recent WCN-based approach, namely WCN-BERT~\cite{Liu2020wcn}. Originally, the WCN-BERT architecture consists of three parts: a BERT encoder for jointly encoding, an utterance representation model, and an output layer for predicting semantic tuples. The BERT encoder exploits posterior probabilities of word candidates in WCNs to inject ASR confidences. Multi-head self-attention is applied over both WCNs and system acts to learn context-aware hidden states. The utterance representation model produces an utterance-level vector by aggregating final hidden vectors. Finally, WCN-BERT adds both discriminative and generative output layers to predict semantic tuples. WCN-BERT stands out for being able to leverage the timing information and confidence scores that are part of standard WCNs. Authors evaluated the performance of WCN-BERT on DSTC2 dataset \cite{henderson2014second}, a corpus of dialogs in the restaurant search domain between a human and automatic agent (i.e., human-machine conversations) reporting  and overall F1=87.91\%.  

For our experiments, we dropped the semantic tuple classifier and the transformer-based hierarchical decoder proposed in the original WCN-BERT paper \cite{Liu2020wcn}. We only preserve the WCN-BERT encoder and the multi-head attention layers to generate the utterance-level representation. On top of this, we concatenate a fully-connected layer to perform intent classification.  

\subsection{Multimodal SLU}
\label{subsec:MulT}


Multimodal SLU refers to the process of embeddings alignment for explicitly minimizing the distance between speech embeddings and the text embeddings from state-of-the-art text encoders like BERT \cite{devlin2019bert}. Thus, the speech embeddings that are used for downstream tasks are made to share a common embedding space with the textual embeddings, leading to better performance in SLU tasks, e.g., intent detection. However, there are a few challenges involved in the process of modeling such multimodal human language time-series, namely: 1) inherent non-aligned data due to variable sampling rates for the sequences from each modality; and 2) long-range dependencies between elements across modalities. 

In order to address these problems, we implemented a solution based on a recent approach named Multimodal Transformer (MulT) \cite{tsai2019multimodal}. MulT depicts an end-to-end model that extends the standard Transformer network \cite{vaswani2017attention} to learn representations directly from unaligned multimodal streams. 
At the heart of MulT, there is a cross-modal attention module, which attends to the crossmodal interactions at the scale of the entire utterances. It merges multimodal time-series via a feed-forward fusion process from multiple directional pairwise crossmodal transformers. Specifically, each crossmodal transformer serves to repeatedly reinforce a target modality with the low-level features from another source modality by learning the attention across the two modalities’ features. 



In our experiments, we adopted the ideas proposed in MulT \cite{tsai2019multimodal}. Hence, given two input modalities, each crossmodal transformer block (one for each modality) keeps updating its sequence. Thus, the crossmodal transformer learns to correlate meaningful elements across modalities. As a final step, outputs are concatenated and passed through a self-attention module to collect temporal information to make predictions. The last elements of the sequence are passed through fully-connected layers to make the intent prediction.

\subsection{XLSR-53: Automatic Speech Recognition Module}
\label{subsect:xlsr}

As shown in Figure \ref{fig:experiments}, we consider as main ASR component the \textit{XLSR-53} pre-trained acoustic model~\cite{conneau2020unsupervised}. 
XLSR-53 learns cross-lingual speech embeddings by pretraining a single generic model from raw waveform of speech in multiple languages. The structure of XLSR is similar to Wav2Vec 2.0~\cite{baevski2020wav2vec}, which is trained using contrastive loss over masked latent speech representations and jointly learns a quantization of the latent embeddings shared across languages. 
XLSR-53 model is trained using 56,000 hours of untranscribed audio data from 53 languages.
We then fine-tune XLSR-53 model~\cite{conneau2020unsupervised} with 390 hours of English data from AMI and Switchboard datasets using E2E-LFMMI loss function~\cite{shao2020pychain, hadian2018end} with biphone units~\cite{vyas2021comparing,wang2019espresso,madikeri2020pkwrap}. 
A grapheme-based lexicon of size 1M was used, and the language model (LM) was trained with 34M utterances from publicly available English datasets including People's speech, Fisher, Switchboard, AMI, Wikitext103, and subsets of Common Crawl and Reddit datasets. 
For improving the generalization of XLSR-53 model to conversational speech in English language, we fine-tuned it using 560 hours of untranscribed data crawled from YouTube. This subset was selected from conversational video calls in English language.
On the YouTube data, we followed an incremental semi-supervised learning approach with four iterations~\cite{khonglah2020incremental}. For decoding, we use the WFST decoder from Kaldi~\cite{povey2011kaldi} toolkit with a beam width of 15.



\begin{table}[t]
    \centering
    \scriptsize
    \caption{SLURP statistics. \textbf{SLURP$_O$} the original dataset, while \textbf{SLURP$_F$} a cleaner version of the original SLURP data.}
    \vspace{0.1cm}
    \begin{tabular}{lll}
        \toprule
        \rowcolor{Gray} \textbf{Statistics} & \textbf{SLURP$_O$} & \textbf{SLURP$_F$} \\
        \midrule
        Audio Files & 72,277 & 50,568 \\
        \quad $\hookrightarrow$ Close range & 34,603 &  25,799 \\
        \quad $\hookrightarrow$ Far range & 37,674 &  24,769 \\
        \cmidrule(lr){1-3}
        Duration [hr] & 58 &  37.2 \\
        Av. length [s] & 2.9 & 2.6 \\
        Nb. of intents & 48 & 47\\
        \bottomrule
    \end{tabular}
    \label{tab:slurp}
\end{table}

\begin{table*}[t]
    \centering    
    \footnotesize
    \caption{Accuracy (ACC) and F1-scores (F1) on intent classification for different representations. We test our approach with either manual or 1-best approaches. Manual refers to ground truth evaluation, while 1-best is obtained by using our ASR module XLSR-53. Thus, \mbox{manual $\rightarrow$ manual} represents the oracle scenario (upper bound), while 1-best $\rightarrow$ 1-best depicts a more real-world case scenario. 
    }
    
    \vspace{0.1cm}
    \begin{tabular}{l l l l c l cccc l cccc}
        \toprule
        \multirow{3}{*}{\textbf{Exp.}} &  &  \multirow{3}{*}{\centering \textbf{Input Type}} &   & \multirow{3}{*}{\textbf{XLSR-53}} & &\multicolumn{4}{c}{\textbf{SLURP$_O$ }} &  & \multicolumn{4}{c}{\textbf{SLURP$_F$ }} \\
        \cmidrule(lr){7-10} \cmidrule(lr){12-15}
         &  &  &  &  &  &\multicolumn{2}{c}{\textbf{Dev ($\uparrow$)}} & \multicolumn{2}{c}{\textbf{Test ($\uparrow$)}} &  & \multicolumn{2}{c}{\textbf{Dev ($\uparrow$)}} & \multicolumn{2}{c}{\textbf{Test ($\uparrow$)}} \\
         \cmidrule(lr){7-8} \cmidrule(lr){9-10} \cmidrule(lr){12-13} \cmidrule(lr){14-15}
         &  & \textit{train}$\rightarrow$\textit{dev-test} &  & \textbf{adaptation} &  & ACC & F1 & ACC & F1 &  & ACC & F1 & ACC & F1 \\
         
        \midrule
        \rowcolor{Gray} \multicolumn{15}{l}{\textbf{\quad \textit{Conventional NLU/SLU}}} \\
        \midrule 
        
        EXP1 &  & manual $\rightarrow$ manual &  &  \textit{NA} &  & \textbf{0.89} & \textbf{0.88} & 0.85 & 0.84 &  & \textbf{0.88 }& \textbf{0.87} & 0.82 & 0.82 \\
        EXP2 &  & manual $\rightarrow$ 1-best &  & \XSolidBrush &  & 0.70 & 0.65 & 0.69 & 0.65 &  & 0.74 & 0.69 & 0.71 & 0.67 \\
        EXP3 &  & 1-best $\rightarrow$ manual &  & \XSolidBrush &  & 0.85 & 0.85 &\textbf{ 0.86} & \textbf{0.85}&  & 0.86 & 0.86 & \textbf{0.85} & \textbf{0.83} \\
        EXP4 &  & 1-best $\rightarrow$ 1-best &  & \XSolidBrush &  & 0.72 & 0.68 & 0.73 & 0.69 &  & 0.76 & 0.71 & 0.77 & 0.73 \\
        \midrule
        EXP5  & &manual $\rightarrow$ 1-best & &\checkmark  & & 0.82 & 0.81 & 0.80 & 0.79 & & 0.84 & 0.82 & 0.86 & 0.84\\
        EXP6  & &1-best $\rightarrow$ manual & &\checkmark  & &\textbf{ 0.88} & \textbf{0.87} & \textbf{0.87} & \textbf{0.86} & &\textbf{ 0.88} & \textbf{0.87} & \textbf{0.88} & \textbf{0.87}\\
        EXP7  & &1-best $\rightarrow$ 1-best & &\checkmark  & & 0.84 & 0.83 & 0.83 & 0.83 & & 0.85 & 0.84 & 0.85 & 0.84 \\

        \midrule
        \rowcolor{Gray} \multicolumn{15}{l}{\textbf{\quad \textit{Lattice-based SLU}}} \\        
        
         
        \midrule
        EXP8  & & WCN & &\XSolidBrush& & 0.68 & 0.67 & 0.68 & 0.68 & & 0.69 & 0.68 & 0.68 & 0.68\\ 
        EXP9  & & WCN  & &\checkmark  & & \textbf{0.78} & \textbf{0.77} & \textbf{0.79} & \textbf{0.79} & & \textbf{0.80} & \textbf{0.80} & \textbf{0.78} & \textbf{0.77}\\        
        \midrule
        \rowcolor{Gray} \multicolumn{15}{l}{\textbf{\quad \textit{Multimodal SLU}}} \\
        \midrule
        
        EXP10 &  & multimodal  &  & \XSolidBrush &  & 0.75 & 0.75 & 0.74 & 0.73 &  & 0.75 & 0.74 & 0.76 & 0.76 \\
        EXP11 &  & multimodal  &  & \checkmark&  & 0.82  & 0.82 & 0.83 & 0.83 &  & 0.83 & 0.83 & 0.82 & 0.82 \\
        EXP12 &  & multimodal {\scriptsize (HuBERT~\cite{hsu2021hubert})} &  & \checkmark&  &  \textbf{0.87} & \textbf{0.88} & \textbf{0.84} & \textbf{0.84} & & \textbf{0.88} & \textbf{0.88} & \textbf{0.86} & \textbf{0.86} \\        
         \bottomrule
    \end{tabular}
    \label{tab:results}
\end{table*}

\section{Experiments and Results}
\label{sect:exps}

\subsection{SLURP Dataset}

To perform our experiments we used the SLURP dataset \cite{bastianelli2020slurp}, a publicly available multi-domain dataset for E2E-SLU, which is substantially bigger and more diverse than other SLU resources. SLURP is a collection of audio recordings of single-turn user interactions with a home assistant. Table~\ref{tab:slurp} contains a few statistics about SLURP.

During a manual analysis, we found many inconsistencies in SLURP annotations. Basically, we identified cases where the manual transcription does not correspond to what is being said in its corresponding audio file. Thus, we considered as erroneous those audio files for which the manual transcription did not match with the automatic transcription, or whose transcripts were inconsistent (i.e., not the same) in the corresponding metadata files. By following this approach, we detected that nearly 30\% (20 hours) of the original data contains some type of inconsistency. We refer as SLURP$_{F}$ to the subset of SLURP without these inconsistent files (see Table~\ref{tab:slurp}).

\subsection{Experiments}
\label{subsect:exps_setup}

Experiments from \textbf{EXP1} - \textbf{EXP7} correspond to the results of conventional NLU/SLU techniques. As described in Section \ref{subsec:pipeline_slu}, these experiments use our implementation of the HERMIT architecture \cite{vanzo-etal-2019-hierarchical}. As can be observed in Table \ref{tab:results}, differences among these experiments are on the type of data used for training and evaluating the HERMIT model, i.e., combinations of either manual transcripts or 1-best ASR outputs. \textbf{EXP8} - \textbf{EXP9} correspond to the experiments done using WCN-based representations (Section \ref{subsec:WCN}). Notice that for both set of experiments, i.e., conventional NLU and WCN-based, we usde the XLS-R model, not-adapted and adapted to SLURP, to obtain the transcripts and WCNs respectively. 
Finally, \textbf{EXP10-EXP12} corresponds to the experiments done using the crossmodal transformer described in Section \ref{subsec:MulT}. Similarly, we evaluate the performance of this approach under circumstances where the XLS-R model is not adapted to the target domain (EXP10), and when adapted to SLURP (EXP11), and one last experiment using acoustic embeddings obtained from HuBERT (EXP12) model.\footnote{\label{note_speechbrain}To generate the acoustic embeddings we followed the \mbox{SpeechBrain}~\cite{speechbrain} SLURP recipe: \url{https://github.com/speechbrain/speechbrain/tree/develop/recipes/SLURP}} 


\subsection{Results}
\label{subsect:results}

\begin{table}
    \centering
    \footnotesize
    \caption{WER\% on SLURP Test sets with the XLSR-53 English model before and after adaptation with SLURP$_F$ train subset.}
    \begin{tabular}{l c c c c}
        \toprule
        System & \multicolumn{2}{c}{Dev (WER\%)} & \multicolumn{2}{c}{Test (WER\%)} \\
        \cmidrule(lr){2-3} \cmidrule(lr){4-5}
               & Headset & All & Headset & all \\
        \midrule
        No adaptation & 23.4 & 34.0 & 23.0 & 34.4\\
        Adapted to SLURP & 13.3  & 16.1 & 13.0 & 15.5 \\
        \bottomrule
    \end{tabular}
    \label{tab:slurp_wer}
\end{table}

Table \ref{tab:results} shows the obtained experimental results for all the benchmarked architectures. Column ``Exp'' indicates the name of the experiment,  ``Input Type'' describes what type of data was used for training and evaluating (dev and test) the corresponding experiment. For those experiments with the tag \textit{manual} it means ground truth transcriptions were used, while \textit{1-best} refers to the automatically generated transcriptions using the XLSR-53 model. Column ``XLSR-53 adaptation'' indicates whether or not the XLSR-53 model was fine-tuned to the SLURP dataset. In order to do the XLSR-53 adaptation, the English ASR model described in Section 2.4 was fine-tuned with the train subset of the SLURP$_F$ data without changing the LM. ASR performances before and after fine-tuning to SLURP are given in Table~\ref{tab:slurp_wer}. And finally, SLURP$_O$ and SLURP$_F$ depict what version of the SLURP dataset was used. 

Notice that the results obtained in the test partition from the cleaned version of the data, i.e., SLURP$_F$, are usually better than those obtained in the original version of the SLURP dataset. To some extent, this is an indicator that the identified inconsistencies in the original SLURP dataset were affecting the benchmarked models, resulting in a miss classification of some intents types.

Experiments EXP1, EXP3, and EXP6 represent artificial scenarios, as in a real-world application we do not expect to have manual transcripts for test partitions. Nevertheless, the best performance under this configuration, e.g., F1=87\% in the SLURP$_F$ for EXP6, represents an upper bound value. Interestingly, this value is even better than the performance obtained in the EXP1, i.e., considering only ground truth data. This may be due to an (unexpected) regularization effect caused by the noise contained in the 1-best transcripts from all the audios of the SLURP.
Thus, it becomes really relevant that WCN-based approaches (EXP8 \& EXP9) are able to obtain a competitive performance against the pipeline \mbox{1-best $\rightarrow$ 1-best} NLU experiments (EXP4). Even though the not-adapted WCN model (EXP8) does not outperform EXP4, this result validates the existence and the impact of richer ASR hypotheses in the lattice, which helps improve the performance of the SLU system, especially in noisy data (SLURP$_{O}$).

Although WCN experiment EXP9 showed a good improvement against the 1-best scenario, multimodal experiments obtain a remarkable performance, comparable to the performance of the oracle experiment (EXP1). The main difference between EXP11 and EXP12 is that the former uses XLSR-53 adapted to SLUPR, while the latter fine-tunes HuBERT toward intent and slot classification in SLURP. This is the explanation for why EXP12 performance is slightly better than EXP11 (XLSR-53 adapted). Finally, as reference results from an E2E approach, the SLURP recipe reports F1 values of F1$=0.77$ and F1$=0.88$ under configurations referred to as \textit{direct} and \textit{direct Hubert} respectively. More details can be found in the respective repository.$^3$ 
Overall, using a multi-modal approach seems to be the recommended option as it guarantees the best performance. Although it should be considered that it represents a costly solution in terms of computational power. On the contrary, if access to manual transcripts is guaranteed for training an NLU/SLU system, independently of having (or not) the possibility to adapt the ASR model toward the target domain, the recommended solution would be to follow a traditional NLU pipeline.

\section{Conclusions}

In this work, we successfully benchmark several neural architectures to perform NLU, pipeline SLU and multi-modal SLU. Our analysis includes SOTA NLU/SLU techniques and compares them in more realistic scenarios. The presented analysis shed light on state-of-the-art architectures in the SLU domain, helping future researchers to define more clearly the application scenario of their proposed solutions.
As an additional contribution, we also put together a cleaner version of the well-known SLURP dataset. During our experimentation process, we found many inconsistencies between the manual annotations and what was really spoken in the audio files. This rise concern on the SLU field, as several papers have already reported results on this dataset without being aware of it. We release the filtered version of SLURP in our git repository, in order to allow others to replicate our experiments.$^2$ 

\bibliographystyle{IEEEtran}
\bibliography{references}
\end{document}